\documentclass[11pt]{article}
\usepackage{fullpage,here,natbib,psfig}

\title{Manual Annotation of Translational Equivalence: \\ The Blinker Project}
\author{I. Dan Melamed \\ Dept. of Computer and Information Science \\
University of Pennsylvania \\ Philadelphia, PA, 19104, U.S.A. \\
{\tt melamed@unagi.cis.upenn.edu} \\
{\tt http://www.cis.upenn.edu/\~{}melamed}
}

\newcommand{\ignore}[1]{}

\date{}

\begin{document}

\maketitle

\section*{Abstract}
Bilingual annotators were paid to link roughly sixteen
thousand corresponding words between on-line versions of the Bible in
modern French and modern English.  These annotations are freely
available to the research community from {\tt
http://www.cis.upenn.edu/\~{}melamed}.  The annotations can be used
for several purposes.  First, they can be used as a standard data set
for developing and testing translation lexicons and statistical
translation models.  Second, researchers in lexical semantics will be
able to mine the annotations for insights about cross-linguistic
lexicalization patterns.  Third, the annotations can be used in
research into certain recently proposed methods for monolingual
word-sense disambiguation.  This paper describes the annotated texts,
the specially-designed annotation tool, and the strategies employed to
increase the consistency of the annotations.  The annotation process
was repeated five times by different annotators.  Inter-annotator
agreement rates indicate that the annotations are reasonably reliable
and that the method is easy to replicate.

\section{Introduction}

% [if time, write about how the data would be used in a test]

Appropriately encoded expert opinions about which parts of a text and
its translation are semantically equivalent can accelerate progress in
several areas of computational linguistics.  First, researchers in
lexical semantics can mine such data for insights about
cross-linguistic lexicalization patterns.  Second, \citet{resyar} have
suggested that cross-linguistic lexicalization patterns are an
excellent criterion for deciding what sense distinctions should be
made by monolingual word-sense disambiguation algorithms.  My own
motivation was in a third area.  Until now, translation lexicons and
statistical translation models have been evaluated either subjectively
\citep[{\em e.g.}\ ][]{arpa} or using only approximate metrics, such
as perplexity with respect to other models \citep{ibm}.  Both
representations of translational equivalence can be tested objectively
and more accurately using a ``gold standard'' such as the one
described here.

Bilingual annotators were paid to link roughly sixteen thousand
corresponding words between on-line versions of the Bible.  As
explained in Section~\ref{gsbitext}, this text was selected to
facilitate widespread use and standardization, which was not a goal or
an outcome of a similar earlier project \citep{sadler}.  A further
distinguishing characteristic of the present work is its emphasis on
measurable consistency.  The annotations were done using the
specially-designed annotation tool described in Section~\ref{blinker},
following a specially-written style guide \citep{styleguide}.
Inter-annotator agreement rates are reported in Section~\ref{IAA}.

\section{The Gold Standard Bitext}
\label{gsbitext}
The first step in creating the gold standard was to choose a bitext.
To make my results easy to replicate, I decided to work with the
Bible.  The Bible is the most widely translated text in the world, and
it exists in electronic form in many languages.  Replication of
experiments with the Bible is facilitated by its canonical
segmentation into verses, which is constant across all
translations\footnote{\citet{tongues} discuss exceptions.}.  After
some simple reformatting, {\em e.g.}\ using the tools described by
\citet{tongues}, the verse segmentation can serve as a ready-made,
indisputable and fairly detailed bitext map.  Among the many languages
in which the Bible is available on-line, I chose to work with two of
the languages with which I have some familiarity: modern French and
modern English.  For modern English I used the New International
Version (NIV) and for modern French the Edition Louis Segond, 1910
(LSG).\footnote{Both are on-line at {\tt
http://bible.gospelcom.net}. Use of the NIV requires a research
license (International Bible Society, Attn: NIV Permission Director,
1820 Jet Stream Drive, Colorado Springs, CO 80921-3969).  LSG is
freely downloadable for research purposes; see {\tt
http://cedric.cnam.fr/ABU/}.}

Once I decided to work with the Bible, I had to decide which parts of
it to annotate.  There is no universal agreement on which books
constitute the Bible, so my decision on which books to include was
guided by two practical considerations.  First, the plurality of
on-line versions includes a particular set of 66 books
\citep{tongues}.  From these 66, I excluded the books of {\em
Ecclesiastes}, {\em Hosea} and {\em Job}, because these books are not
very well understood, so their translations are often extremely
inconsistent \citep{aster}.  The remaining 63 books comprise 29614
verses.  My choice of verses among these 29614 was motivated by the
desire to make the gold standard useful for evaluating
non-probabilistic translation lexicons.  The accuracy of an
automatically induced translation lexicon can be evaluated only in
terms of the bitext from which it was induced \citep{amta}: Reliable
evaluation of a word's entry in the lexicon requires knowledge of all
of that word's translations in the bitext.  Therefore, I decided to
annotate a set of verses that includes all instances of a set of
randomly selected word types.  However, the set of word types was not
completely random, because I also wanted to make the gold standard
useful for investigating the effect of word frequency on the accuracy
of translation lexicon construction methods.

To meet this condition, I used the following procedure to select
verses that contain a random sample of word types, stratified by word
frequency.
\begin{enumerate}
\item I pre-processed both halves of the Bible bitext, to separate
punctuation symbols from the words to which they were adjacent and to
split elided forms (hyphenated words, contractions, French {\em du}
and {\em aux}, {\em etc.}) into multiple tokens.  To keep the bitext
easy to read, I did not lemmatize inflected forms.  The resulting
bitext comprised 814451 tokens in the English half and 896717 tokens
in the French half, of 14817 and 21372 types, respectively.
\item I computed a histogram of the words in the English Bible.
\item I randomly selected a {\bf focus set} of one hundred word types,
consisting of twenty-five types that occurred only once, twenty-five
types that occurred twice, twenty-five types that occurred three times
and twenty-five types that occurred four times.
\item I extracted the English verses containing all the instances of
all the words in the focus set, and the French translations of those
verses.
\item Step~4 resulted in some verses being selected more than once,
because they contained more than one of the words in the focus set.  I
eliminated the duplications by discarding the lower-frequency word in
each conflict and resampling from the word types with that frequency.
\end{enumerate}
The one hundred word types in the final focus set are listed in
Table~\ref{focusset}.  The tokens of these word types are contained in
$(1 + 2 + 3 + 4) * 25 = 250$ verse pairs.  By design, all the possible
correct translations of the focus words in the bitext can be
automatically extracted from the annotations of these 250 verse pairs.
Including the focus words, the 250 verses in the gold standard
comprise 7510 English word tokens and 8191 French word tokens, of 1714
and 1912 distinct types, respectively.

\begin{table*}[htb]
\centering
\begin{tabular}{|l|l|l|l|}
\hline
Frequency 1 & Frequency 2 & Frequency 3 & Frequency 4 \\ \hline \hline

Akkad & Alexandrian & Beginning & Anointed \\
Arnan & Around & Cover & Derbe \\
Ashterathite & Carites & Formerly & Izharites \\
Bimhal & Dressed & Gatam & Jeriah \\
Cun & Exalt & Inquire & Mikloth \\
Ephai & Finish & agrees & assurance \\
Ethiopians & Halak & deceivers & burnished \\
Harnepher & Helam & defended & circle \\
Impress & Jahleel & defiling & defender \\
Jairite & Jokmeam & drain & dens \\
Jeberekiah & Kehelathah & engulfed & examined \\
Manaen & Plague & equity & failing \\
Nekeb & Zeus & evident & herald \\
apt & brandish & goldsmiths & leadership \\
eyesight & fulfilling & intense & loathe \\
handmill & hotly & partners & radiance \\
improperly & intelligible & profound & rallied \\
journeys & ledges & progress & refusing \\
origins & lit & rout & secretaries \\
parade & pardoned & stared & student \\
readily & petitioned & starting & stumbles \\
unending & reappears & swirling & thankful \\
unsatisfied & thwarts & thistles & topaz \\
unsuited & undoing & tingle & violently \\
visitors & unscathed & woodcutters & wisely \\

\hline
\end{tabular}
\caption{{\em Word types in the gold standard's focus
set.}\label{focusset}}
\end{table*}

\section{The Blinker Annotation Tool}
\label{blinker}

\begin{figure}[H]
\centerline{\psfig{figure=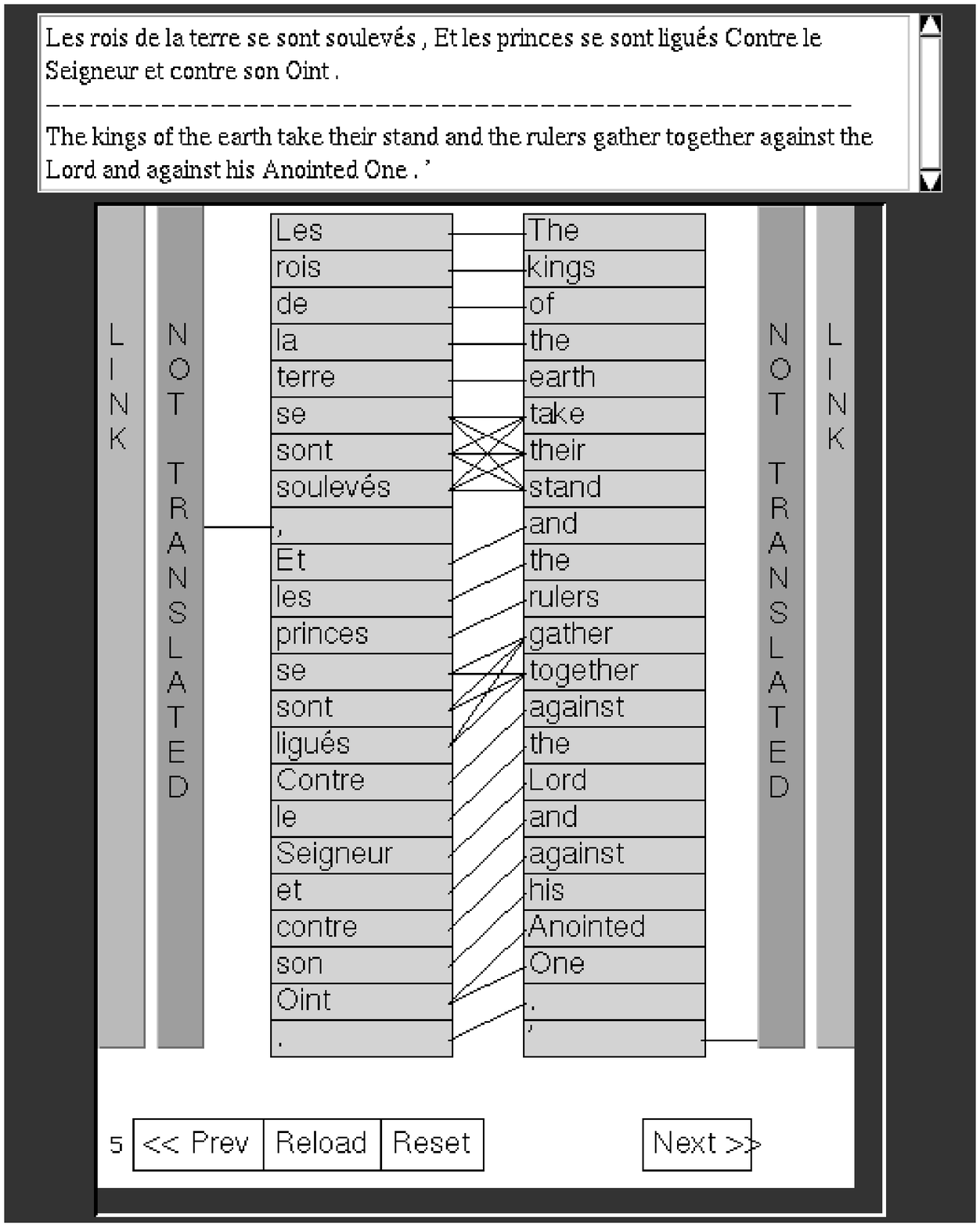,width=5.93in}}
\caption{{\em A Blinker session.}\label{fig:blinker}}
\end{figure}

To promote consistent annotation, I needed an effective way to link
corresponding words in the bitext. I could have just asked
bilingual annotators to type in pairs of numbers corresponding to the
word positions of mutual translations in their respective verses.
However, such a data entry process would be so error-prone that it
would render the annotations completely unreliable.  Instead, I
designed the Blinker (``bilingual linker''), a mouse-driven graphical
annotation tool.  The Blinker was implemented at the University of
Maryland, under the direction of Philip Resnik.  The Blinker makes
heavy use of color-coding, but a greyscale screen capture of a Blinker
session is shown in Figure~\ref{fig:blinker}.  To get an idea of how
the Blinker was used, refer to the instructions in
Figure~\ref{blink_instr}.
\begin{figure}[H]
% \begin{singlespace}
\begin{tabular*}{6.5in}{c}
\hline
\end{tabular*}
\centerline{\Large How to Use the Blinker}
\vspace*{.1in}

\hspace*{.25in} The Blinker (for ``bilingual linker'') is a mouse-driven graphical user interface.  Here's how to use it:
\begin{itemize}
\item
To specify the correspondence between two or more words,
\begin{enumerate}
\item Select the words you want by clicking on them with the LEFT
mouse button.  The boxes around the words will turn pink.  (Note:
Clicking on a word again will ``unselect'' it.)  
\item Either click on one of the \fbox{Link} bars, or click the MIDDLE or
RIGHT mouse button. The Blinker will draw lines between the words that
you've selected and color their boxes light blue.
\end{enumerate}
\item
To specify that a word is not translated, click on the word (it will
turn pink), and then click on the \fbox{Not Translated} bar beside it.
The Blinker will draw a line from the word to the \fbox{Not Translated}
bar.
\item If you ever change your mind, you can simply re-link words that
you've already linked.  The Blinker will delete all the links
previously associated with those words and draw the new links that
you've specified.
\item You will see four buttons at the bottom of each verse pair.
\begin{enumerate}
\item
When you've finished specifying all the correspondences for a pair of
verses, click the \fbox{Next $>>$} button at the bottom.  The Blinker will
verify that all the words on both sides have been accounted for, and
then present you with the next pair of verses in the set.
\item
The \fbox{$<<$ Prev} button allows you to return to previous verse pairs
in the same set.
\item The \fbox{Reset} button allows you to erase all the links in the
verse pair currently on the screen.
\item The \fbox{Reload} button allows you to reload the most recent links
for the whole set of verse pairs, e.g. after you've pressed Reset or
when you return to a set after taking a break.
\end{enumerate}
{\bf Your work is permanently saved whenever you press \fbox{Next $>>$},
\fbox{$<<$ Prev} or \fbox{Reload}.}
\end{itemize}
\begin{tabular*}{6.5in}{c}
\hline
\end{tabular*}
% \end{singlespace}
\caption{{\em Blinker instructions.}\label{blink_instr}}
\end{figure}

\section{Methods for Increasing Reliability}

Translational equivalence is often difficult to determine at the word
level.  Many words are translated only as part of larger text units,
and even Biblical translations are sometimes inconsistent or
incomplete.  Therefore, I adopted several measures to increase the
reliability of the gold standard annotations.

First, instead of relying on only one or two annotators, I recruited
as many as I could find --- seven --- with the intent of creating
multiple annotations for the same data.  Each set of annotations could
be compared to the others, in order to identify deviations from the
norm.  The replication of effort also enabled evaluation of the gold
standard itself in terms of inter-annotator agreement rates.

Second, I designed the Blinker to prevent an annotator from proceeding
to the next verse pair until all the words in the current verse pair
were annotated.  If an annotator felt that a given word did not have a
translational equivalent in the opposite verse, she had to explicitly
mark the word as ``Not Translated.''  This forced-choice annotation
method can be contrasted with the strategy adopted in the Penn
Treebank project for part-of-speech (POS) annotation \citep{treebank}.
Most of the Penn Treebank was annotated for POS only once, and the
annotation method was to manually correct the output of an automatic
POS tagger.  \citet{treebank} have reported that this method produced
more reliable annotations than manual POS tagging from scratch.
However, a reliable ``corrective'' annotation method is only possible
given a reasonably good first approximation.  Such an approximation
might have been achieved by one of the translation models described in
the literature \citep[{\em e.g.},][]{ibm,transmod}, but only at the
expense of biasing the gold standard towards a particular translation
model, which would have defeated the purpose of the project.  Another
justification for the forced-choice approach is the overwhelming bias
towards a ``no~link'' annotation --- the vast majority of word pairs
are not linked.  When I attempted the task myself, I was amazed at how
many words I accidentally neglected to annotate.  A disadvantage of
the forced-choice approach is that forced decisions are not reliable
when they are difficult.  The reliability of the gold standard will be
measured in Section~\ref{IAA}.

My third strategy for increasing the reliability of the gold standard
was borrowed from the Penn Treebank project \citep{treebank}: I
constructed an annotation style guide \citep{styleguide}.  To reduce
experimenter bias, the guide was based largely on the intuitions of
the annotators:
\begin{enumerate}
\item I wrote a draft version of the General Guidelines.
\item Two groups of annotators each annotated a set of ten randomly
selected verse pairs from the Bible bitext, using the draft General
Guidelines.  There were seven annotators, so one set of ten verse pairs
was annotated four times and the other three times.  These dry-run
annotations also served the purpose of acclimating the annotators to
the task.
\item The different annotations for each verse pair were
automatically compared.
\item I manually analyzed the differences and identified the major
sources of variation in the annotations.
\item I reconvened four of the seven annotators, and presented them
with examples of the different kinds of variation, one kind at a time.
We briefly discussed each kind of variation, and then the annotators
voted on the preferred annotation style.
\item I compiled the votes and the examples on which they were based
into the Detailed Guidelines \citep[see][for details]{styleguide}.  I
also added some clarifying examples post-hoc.
\item When the annotators began annotating the gold standard, they
reported a few additional difficult cases.  I solicited votes on the
preferred annotation style for these difficult cases from all the
annotators by email.  The majority opinions were incorporated into the
style guide. 
\end{enumerate} \nopagebreak
The annotators were encouraged to conform to the style guide by a
financial incentive plan, which was informally described to them in
the message in Figure~\ref{bonusplan}. The description of the
incentive plan was intentionally vague, to prevent any attempts to
game the system.  The annotators' base pay rate was set by my
university at either \$8.50 or \$10.80 per hour, depending on whether
they had finished college.  So, a \$200 bonus would have seemed
substantial.
\begin{figure}[H]
% \begin{singlespace}
\begin{tabular*}{5.93in}{c}
\hline
\end{tabular*}
\begin{verbatim}
As I mentioned at the kick-off meeting, we're more interested in
consistency than correctness of annotations -- correctness is often
quite subjective for this task.  We are investing alot of effort into
getting highly consistent annotations; the success of our project
depends on it.  Improving consistency is the whole reason behind
creating a style guide.  Since we've gone this far, we're willing to
go a little further, and offer you some financial incentive to
carefully follow the style guide.

Here's how it will work.  For each "difficult" verse pair, we will
compute a "link correlation" matrix among all the annotators who
worked on that verse pair.  E.g. if the annotators are A1 through A5,
we might end up with a matrix like this:

        A2      A3      A4      A5      Average

A1      23      35      43      34      35
A2              65      45      85      65
A3                      76      45      56
A4                              45      85
A5                                      53

(The numbers are in %; I didn't actually calculate the averages, I
just typed in some random numbers.)

Within each set, the two annotators with the lowest average
correlation get no bonus.  The annotator with the third highest
average correlation gets a bonus of X. The annotator with the second
highest average correlation gets a bonus of 2X. The annotator with the
highest average correlation gets a bonus of 5X.  X is determined by
the number of sets we end up with, but the total bonus pool is $320.
Thus, if you closely follow the style guide, you can score up to $200
extra.

I realize this is a little convoluted.  Please let me know if you have
questions.
\end{verbatim}
\begin{tabular*}{5.93in}{c}
\hline
\end{tabular*}
% \end{singlespace}
\caption[{\em Financial incentive plan for annotator
consistency}]{{\em The intentionally vague financial incentive to
conform to the annotation style guide, emailed to all
annotators.}\label{bonusplan}}
\end{figure}

\section{The Annotators}

Before the kick-off meeting, all annotators answered a brief
questionnaire, which included some administrivia followed by the
questions in Table~\ref{questionnaire}.
\begin{table}[htb]
% \begin{singlespace}
\begin{tabular*}{5.93in}{c}
\hline
\end{tabular*}
\begin{verbatim}
Q3:  Have you ever taken a course in syntax?
     (of the kind taught in linguistics departments)

Q4:  Please rate your level of fluency in French:
        a) French linguist or professional translator
        b) native speaker
        c) near-native, due to, e.g., long-term residence in France
        d) proficient enough to satisfy Penn's foreign language requirement
        e) took it in high-school

Q5:  Please rate your level of fluency in English:
        a) English linguist or professional translator
        b) native speaker
        c) proficient enough to ace TOEFL
        d) I plan to work through an interpreter

Q6:  Given that the project may require up to 20 hours of your time, how
        long do you think it will take you to finish?
        a) one week
        b) two weeks
        c) three weeks
        d) a month
        e) longer
\end{verbatim}
\begin{tabular*}{5.93in}{c}
\hline
\end{tabular*}
% \end{singlespace}
\caption{{\em Part of the questionnaire given to
annotators.}\label{questionnaire}}
\end{table}
\noindent ``Penn's foreign language requirement'' in Q4 refers to the
University of Pennsylvania's policy that every undergraduate must be
fluent in a foreign language to get their degree.  ``TOEFL'' in Q5
stands for the Test of English as a Foreign Language that all foreign
students must pass in order to be admitted into the University of
Pennsylvania.  Table~\ref{anns} lists the annotators' responses to
these questions, along with some of their other attributes that I
learned through my personal interaction with them.
\begin{table*}[htb]
\centering
\begin{tabular}{|c||c|c|c|c|c|c|c|}
\hline
Annotator &	Approximate	&	Sex &	Level of&	Q3 &	Q4 & 	Q5 &	Q6 \\	
Code	&	Age	&	&	Education &	&	&	&	\\
\hline \hline
A1	&	30	&	F & 	MA	  &	N &	a &	b &	b \\
A2	&	24	&	M &	almost BSc &	N &	c &	b &	c \\
A3	&	28	&	F &	almost MSc &	Y &	d &	b &	c \\
A4	&	over 60	&	M &	BA	   &	N &	d &	b &	c \\
A5	&	24	&	F &	BA	   &	Y &	b &	c &	c \\
A6	&	21	&	M &	almost BSc &	N &	d & 	b &	b \\
A7	&	21	&	F &	almost BA  &	N &	c-d &	b &	b \\
\hline
\end{tabular}
\caption{{\em The annotators and their responses to the questionnaire}.\label{anns}}
\end{table*}

\section{Inter-Annotator Agreement}
\label{IAA}

The seven annotators annotated the 250 verse pairs five times.  The
distribution of verse pairs among annotators was dictated by how much
time each annotator could devote to the project, as indicated by their
answer to Q6 in the questionnaire.  Table~\ref{verse-distrib} shows
which annotator annotated which verse pairs and how long they took.  I
shall report separate inter-annotator agreement statistics for the two
parts of the gold standard defined in Table~\ref{verse-distrib}.
\begin{table*}[htb]
\centering
\begin{tabular}{|c||r||c|c|c|c|c|}
\hline
& Annotation \# &	1	& 2	& 3	& 4	& 5 \\ \hline \hline
Part 1: verse pairs 1 to 100 & Annotator & A1 & A2 & A3 & A4 & A5 \\
	&	hours spent & 9.5 & 10 & 9 & 10.7 & 10.5 \\  \hline
Part 2: verse pairs 101 to 250 & Annotator & A1 & A2 & A3 & A6 & A7 \\
	&		hours spent & 12 & 18.5 & 11.5 & 22 & 20 \\
\hline
\end{tabular}
\caption{{\em Which annotators annotated which verse
pairs and how long they took.}\label{verse-distrib}}
\end{table*}

The simplest way to measure agreement would have been to compute a
single rate for each pair of annotators over whichever parts of the
gold standard they both annotated.  However, standard deviations could
not be computed this way.  The next simplest way to measure agreement
would have been to compute separate agreement rates for each of the
250 verse pairs, and then to find the means and standard deviations of
these 250 rates.  However, this approach would have resulted in
inflated agreement rates.  The problem was that links in shorter verse
pairs were easier to assign and therefore less likely to diverge.
Since there were fewer links in shorter verse pairs, each of these
``easier'' links would have influenced the mean agreement rate more
than the links in long verse pairs.  A more accurate method for
measuring agreement lay in between the two extremes.  I divided Part~1
of the gold standard into 10 sets of 10 verse pairs each, and Part~2
into 10 sets of 15 verse pairs each.  I pooled the links in each set
of verses and computed 10 agreement rates for each pair of annotators
for each part of the gold standard.  Then, I computed the means and
standard deviations of the 10 rates for each pair of annotators for
each part of the gold standard.

A straightforward metric for measuring agreement rates can be derived
from the recall and precision measures widely used in the information
retrieval literature.  When comparing a set of ``test'' elements $X$
to a set of ``correct'' elements~$Y$,
\begin{equation}
\label{precision}
precision(X | Y) = \frac{|X \cap Y|}{|X|} ,
\end{equation}
\begin{equation}
\label{recall}
recall(X | Y) = \frac{|X \cap Y|}{|Y|} .
\end{equation}
$X$ and $Y$ can be fuzzy sets, such as probability distributions, in
which case $|X|$ is defined as the sum of the weights of the elements
in $X$ and $|X \cap Y|$ is the sum of the weights of the elements
shared by $X$ and $Y$.  Equations~\ref{precision} and~\ref{recall}
differ only in the set whose size is used as the denominator.  If
neither $X$ nor $Y$ is privileged, or if precision and recall are
equally important, we can compute a symmetric measure of
agreement $D$ as the harmonic mean of precision and recall:
\begin{equation}
\label{setDice}
D(X,Y) = \frac{1}{\frac{1}{Precision(X | Y)} + \frac{1}{Recall(X |
Y)}} = \frac{2 * |X \cap Y|}{|X| + |Y|} .
\end{equation}
$D$ is the set-theoretic equivalent of the Dice coefficient
\citep{dice} and conveniently ranges from zero to one.  

From an information-processing point of view, the input to the
annotators was a set of aligned text segments and their output was a
set of pairs of corresponding word positions.  So, inter-annotator
agreement should be measured in terms of the similarity between sets
of pairs of
% [!!!! when there's time, do it by type on the focus set too]
corresponding word positions.  There is a small problem with counting
pairs of word positions at face value, however.  The annotators of the
gold standard could link each word to as many other words as they
wished ({\em e.g.}\ ``take their stand'' in Figure~\ref{fig:blinker}).
Therefore, an evaluation metric that treats all link tokens as equally
important would place undue importance on words that were linked more
than once.

One solution to this problem is to attach a weight $w(u, v)$ to each
link token $(u, v)$, where
\begin{equation}
\label{fanout}
w(u, v) = \frac{1}{\max[fanout(u), fanout(v)]} .
\end{equation}
The $fanout$ function returns the number of links attached to its
argument.  When the link tokens are weighted in this fashion, the
weights attached to each word will sum to at most one.  With the link
weights in place, we can compute precision, recall and $D$ as defined
above.  This solution is mildly deficient, because when the lowest
common multiple of $fanout(u)$ and $fanout(v)$ is neither $fanout(u)$
nor $fanout(v)$, then neither $u$ nor $v$ will carry full weight.
However, such cases are so rare that the problem can be ignored for
the sake of a simple evaluation method.  Some of the evaluations in
the following chapters are based on Equation~\ref{setDice}, weighted
by Equation~\ref{fanout}.

For the purposes of evaluating the gold standard itself, I used a
slightly more complicated but non-deficient weighting scheme.  First,
links were treated as directed pointers from the French side of the bitext
to the English side.  Weights were normalized so that the weights of
the links emitted from any single French word token summed to 1.
However, no limit was placed on the total weight of links that could
point to an English word token.  With the links weighted in this
fashion, an agreement rate $D_{F\rightarrow E}$ was computed between
each pair of annotators using Equation~\ref{setDice}.  Then, the links
were reversed and reweighted so that the weights of the links emitted
from any one English word summed to 1, but the weight of links
pointing to a French word was unrestricted.  A second agreement rate
$D_{E\rightarrow F}$ was computed between each pair of annotators with
the links normalized in this direction.  The final agreement rate was
the mean of $D_{F\rightarrow E}$ and $D_{E\rightarrow F}$.  The rates
for each pair of annotators, for each part of the gold standard, along
with the mean for each annotator and the grand mean are shown in
Table~\ref{iaa}.
\begin{table*}[htb]
\centering

Part 1:  Verse pairs 1-100

\begin{tabular}{|c|c|c|c||c||c|}
\hline
A2 		& A3 		& A4 		& A5 		& annotator & mean \\ \hline \hline
81.81 $\pm$ 4.61 & 89.64 $\pm$ 5.38 & 82.91 $\pm$ 4.73 & 86.06 $\pm$ 4.21 &  A1 & 85.11 $\pm$ 5.67 \\ \hline
   		& 81.71 $\pm$ 3.10 & 79.27 $\pm$ 3.14 & 81.73 $\pm$ 2.75 & A2  & 81.13 $\pm$ 3.71 \\ \hline
   		&		& 82.53 $\pm$ 5.23 & 85.96 $\pm$ 3.11 & A3  & 84.96 $\pm$ 5.38 \\ \hline
   		&  		&  		& 79.54 $\pm$ 3.84 & A4  & 81.06 $\pm$ 4.68 \\ \hline
   		&  		&  		& 		 & A5 & 83.32 $\pm$ 4.53 \\ \hline
		&		&		&		& grand mean & 83.12 $\pm$ 5.16 \\ \hline
\end{tabular}

Part 2:  Verse pairs 101-250

\begin{tabular}{|c|c|c|c||c||c|} 
\hline
A2 		& A3 		& A6		& A7 		& annotator & mean \\ \hline \hline
81.92 $\pm$ 3.97 & 87.85 $\pm$ 2.79 & 77.04 $\pm$ 2.99 & 85.82 $\pm$ 2.02 & A1 & 83.15 $\pm$ 5.12 \\ \hline
		& 81.45 $\pm$ 3.91 & 74.20 $\pm$ 4.11 & 80.50 $\pm$ 3.55 & A2 & 79.52 $\pm$ 4.99 \\ \hline
   		&		& 76.81 $\pm$ 2.89 & 85.00 $\pm$ 2.12 & A3 & 82.78 $\pm$ 5.11 \\ \hline
   		&  		&  		& 75.63 $\pm$ 2.51 & A6 & 75.92  $\pm$ 3.38 \\ \hline
   		&  		&  		& 		 & A7 & 81.74 $\pm$ 4.84 \\ \hline
		&		&		&		& grand mean & 80.62 $\pm$ 5.44 \\ \hline
\end{tabular}

\caption[{\em Inter-annotator agreement on all words}]{{\em Percent
inter-annotator agreement, \mbox{$\pm$ standard deviation}.}
\label{iaa}}
\end{table*}

\ignore{ 
The annotators needed to meet only for the kick-off meeting where the
style guide was created.  In order to ensure that the five annotations
were done independently, they were instructed to work alone
thereafter.  However, the financial incentive described in
Figure~\ref{bonusplan} might have prompted two or more of the
annotators to collaborate.  Fortunately, none of the agreement rates
in Table~\ref{iaa} are much higher than the others (annotator A6 is a
{\em low} outlier), so it is unlikely that any of the annotations are
interdependent.
}

Regardless of how literal the translation is in a given bitext, some
words will not correspond well to words on the other
side.  In particular, the translations of function words often depend
more strongly on the content words around them than on the function
words themselves.  Function words are the first to change when a
translator decides to paraphrase.  Most of the annotation style guide
was devoted to annotation conventions for function words.  These
observations suggest that the inter-annotator agreement may be
higher for content words than for function words.  

Since function words are not important for some applications of
translation models, it is useful to measure the inter-annotator
agreement rates for content words only.  I compiled a stoplist of 287
function words for English and 375 function words for French.  These
lists consisted of all words that were not nouns, verbs, adverbs or
adjectives, in addition to all inflections of all auxiliary verbs
({\em do, go, be, etc.}\ and their French equivalents).  The gold
standard contained 2871 English word tokens and 2768 French word
tokens that were not on the stoplist.  From the complete set of
annotations, I removed all the links that had a stoplisted word on
either side.  Then, I re-evaluated inter-annotator agreement, using
the same method, but only on the remaining links.
Table~\ref{iaa-content} shows the results.  The effect of ignoring
function words is well illustrated by the 10\% rise in the grand mean
of Table~\ref{iaa-content} over the grand mean of Table~\ref{iaa}.

\begin{table*}[htb]
\centering

Part 1:  Verse pairs 1-100

\begin{tabular}{|c|c|c|c||c||c|}
\hline
A2 		& A3 		& A4 		& A5 		& annotator & mean \\ \hline \hline
90.60 $\pm$ 4.62 & 94.37 $\pm$ 4.99 & 91.75 $\pm$ 3.38 & 94.20 $\pm$ 3.28 & A1 & 92.73 $\pm$ 4.45 \\ \hline
		& 90.20 $\pm$ 3.20 & 90.52 $\pm$ 2.94 & 90.54 $\pm$ 2.24 & A2 & 90.46 $\pm$ 3.38 \\ \hline
		& 		& 91.85 $\pm$ 4.69 & 94.33 $\pm$ 3.74 & A3 & 92.69 $\pm$ 4.58 \\ \hline
		&		&		& 92.17 $\pm$ 2.48 & A4 & 91.57 $\pm$ 3.55 \\ \hline
		&		&		&		& A5 & 92.81 $\pm$ 3.40 \\ \hline
		&		&		&		& grand mean & 92.05 $\pm$ 4.01 \\ \hline
\end{tabular}

Part 2:  Verse pairs 101-250

\begin{tabular}{|c|c|c|c||c||c|} 
\hline
A2 		& A3 		& A6		& A7 		& annotator & mean \\ \hline \hline
90.91 $\pm$ 3.81 & 94.17 $\pm$ 2.69 & 88.38 $\pm$ 3.56 & 94.37 $\pm$ 2.57 & A1 & 91.96 $\pm$ 4.06 \\ \hline
		& 90.92 $\pm$ 3.43 & 87.80 $\pm$ 4.20 & 90.79 $\pm$ 3.24 & A2 & 90.11 $\pm$ 3.93 \\ \hline
		&		& 88.88 $\pm$ 4.23 & 93.52 $\pm$ 2.56 & A3 & 91.87 $\pm$ 3.92 \\ \hline
		&		&		& 88.04 $\pm$ 3.36 & A6 & 88.28 $\pm$ 3.90 \\ \hline
		&		&		&		& A7 & 91.68 $\pm$ 3.87 \\ \hline
		&		&		&		& grand mean & 90.78  $\pm$ 4.18 \\ \hline
\end{tabular}

\caption[{\em Inter-annotator agreement on content words only}]{{\em
Percent inter-annotator agreement on content words only, \mbox{$\pm$
standard deviation}.}
\label{iaa-content}}
\end{table*}

\ignore{[!!!! Is this really necessary?  it's not reasonable to suspect
that the annotators were behaving randomly or as a baseline.  Do it if
there's time.]
Table~\ref{iaa} also compares these agreement rates to two different
baseline rates.  The first baseline is obtained by linking each word
position in the shorter of a pair of verses to the same position in
the longer verse.  The second baseline is obtained by linking each
position $p$ in the shorter verse to position $q$ in the longer verse,
where $q = \frac{m}{n} p$ and $m$ and $n$ are the lengths of the
longer and shorter verses, respectively, and $q$ is rounded to the
nearest integer.  }

The inter-annotator agreement rates in Tables~\ref{iaa}
and~\ref{iaa-content} indicate that the annotators were doing mostly
the same thing most of the time, and that the task is reasonably
well-defined and reasonably easy to replicate.  This claim is
strengthened by the observation that annotator A6 was a low outlier
regardless of whether function word links are considered, which is why
the grand means are lower for Part~2 than for Part~1.  Nevertheless,
the agreement rates are not as high as one might like.
Although much more research is required to draw any conclusions
with certainty, I can suggest three reasons why the inter-annotator
agreement rates are not any higher.

First, despite the care taken with Biblical translations, many aligned
Bible verses carry significantly different meanings.  For example:
\begin{quote}
\begin{itemize}
\item[{\bf English:}] They also brought to the proper place their
quotas of barley and straw for the chariot horses and the other
horses.
\item[{\bf French:}] 
Ils faisaient aussi venir de l'orge et de la paille pour les chevaux
et les coursiers dans le lieu o\`{u} se trouvait le roi, chacun selon
les ordres qu'il avait re\c{c}us.
\end{itemize}
\end{quote}
One possible explanation for the divergence is that
neither of my Bible versions is a translation of the
other; rather, both are probably translations of a third original, if
not two different originals.  Furthermore, careful translation
usually does not imply literal translation.  The distinction is
particularly apparent in the case of the Bible.

Second, the style guide was based on only a small sample of annotated
bitext, and it was inevitable that new sources of variation in the
annotations would occur in previously unseen bitext.  In order to
further standardize the annotation style, it would have been necessary
to update the style guide in an iterative manner, with each new batch
of annotated verses being checked for new sources of inter-annotator
variation.  Such a procedure was beyond my time and budget
constraints.

Third, as with all first versions of such tools, the Blinker
annotation tool left much to be desired.  For example, when one of a
pair of verses was significantly longer than the other, the lines
representing some of the links were nearly vertical and blended
together.  One annotator admitted by email, `I do at times throw up my
hands in frustration at how hard it is \ldots to link a word at the
top to a word at the veeeeeeeery bottom.  I reckon you just may get
extra ``not-linked''s because of this.'  A better Blinker design may
have made it easier for the annotators to follow the style guide.

\section{Conclusion}

This chapter described a method for manually constructing explicit
representations of translational equivalence.  After a special
annotation tool was implemented, the method was used to annotate
corresponding words in a significantly large part of a widely
available bitext.  The annotation is intended for use as a gold
standard for comparing automatically constructed models of
translational equivalence, and thus also for comparing the methods
used to construct such models.  Inter-annotator agreement rates on the
gold standard are roughly 82\%, or roughly 92\% if function words are
ignored.  These rates indicate that the gold standard is reasonably
reliable and that the task is reasonably easy to replicate.

\section*{Acknowledgements}
Special thanks to Alex Garthwaite, Robert MacIntyre and Philip Resnik
for advice on the design of the Blinker and to Galen Wilkerson for
much of its implementation.  Thanks also to the annotators, without
whose dedication and invaluable feedback this project would not have
been possible.  And of course thanks to DARPA, without whose grant
N6600194C-6043 this project would not have been possible either.

\clearpage

%%%%%%%%%%%%%%%%%% BIBLIOGRAPHY

\end{document}